\def\year{2019}\relax
\documentclass[letterpaper]{article} 
\usepackage{aaai19}  
\usepackage{times}  
\usepackage{helvet}  
\usepackage{courier}  
\usepackage{url}  
\usepackage{graphicx}  
\usepackage{subfigure}
\usepackage{amsmath}
\usepackage{amsfonts}
\usepackage{multirow}
\usepackage{arydshln}
\usepackage{balance}

\begin{document}
%
\title{DTMT: A Novel Deep Transition Architecture for Neural Machine Translation}
\author{
Fandong Meng and Jinchao Zhang \\
WeChat AI - Pattern Recognition Center  Tencent Inc. \\
\{fandongmeng, dayerzhang\}@tencent.com
}

\maketitle
\begin{abstract}
Past years have witnessed rapid developments in Neural Machine Translation (NMT). Most recently, with advanced modeling and training techniques, the RNN-based NMT (RNMT) has shown its potential strength, even compared with the well-known Transformer (self-attentional) model. Although the RNMT model can possess very deep architectures through stacking layers, the transition depth between consecutive hidden states along the sequential axis is still shallow. In this paper, we further enhance the RNN-based NMT through increasing the transition depth between consecutive hidden states and build a novel Deep Transition RNN-based Architecture for Neural Machine Translation, named \textsc{DTMT}. This model enhances the hidden-to-hidden transition with multiple non-linear transformations, as well as maintains a linear transformation path throughout this deep transition by the well-designed linear transformation mechanism to alleviate the gradient vanishing problem. Experiments show that with the specially designed deep transition modules, our \textsc{DTMT} can achieve remarkable improvements on translation quality. Experimental results on Chinese$\Rightarrow$English translation task show that \textsc{DTMT} can outperform the Transformer model by +2.09 BLEU points and achieve the best results ever reported in the same dataset. On WMT14 English$\Rightarrow$German and English$\Rightarrow$French translation tasks, \textsc{DTMT}\footnote{We release the source code at: https://github.com/fandongmeng/DTMT\_InDec} shows superior quality to the state-of-the-art NMT systems, including the Transformer and the RNMT+.
\end{abstract}

\section{Introduction}
Neural Machine Translation (NMT) with an encoder-decoder~\cite{ChoEMNLP,googleS2S} framework has made promising progress in recent years. Generally, this kind of framework consists of two components: an encoder network that encodes the input sentence into a sequence of distributed representations, based on which a decoder network generates the translation with an attention mechanism~\cite{cho,Luong:15}. Driven by the breakthrough achieved in computer vision~\cite{he2016deep}, research in NMT have turned towards studying deep architectures~\cite{Wu:16,ZhouCWLX16,Kalchbrenner:16,wangEtAl2017,Gehring:17,VaswaniEtal2017}. Among these studies, RNN-based NMT (RNMT) with deep stacked architectures~\cite{Wu:16,ZhouCWLX16,Kalchbrenner:16,wangEtAl2017} first outperforms the conventional shallow RNMT model, to be the de-facto standard for NMT. Most recently, after absorbing the advanced modeling and training techniques, the RNMT+~\cite{ChenACL2018} has shown its greater potential strength, even surpasses the convolutional seq2seq (ConvS2S) model~\cite{Gehring:17} and achieves comparable results with the well-known Transformer model~\cite{VaswaniEtal2017}. These studies inspire researchers to make efforts for searching new architectures for RNMT.

Although the RNMT model can possess very deep architectures through stacking layers, for each recurrent level of the stacked RNN, the transition between the consecutive hidden states along the sequential axis is still shallow. Since the state transition between the consecutive hidden states effectively adds a new input to the summary of the previous inputs represented by the hidden state, this procedure of constructing a new summary from the combination of the previous one and the new input should be highly nonlinear, to allow the hidden state to rapidly adapt to quickly changing modes of the input while still preserving a useful summary of the past~\cite{pascanu2013construct}. From this perspective, some researchers~\cite{pascanu2013construct,barone2017deep} investigate deep transition recurrent architectures, which increase the depth of the hidden-to-hidden transition. This kind of structure extends the conventional shallow RNN in another aspect different from the stacked RNN, and has been proven to outperform the stacked one on language modeling task~\cite{pascanu2013construct}. \citeauthor{barone2017deep}~\shortcite{barone2017deep} apply this transition architecture to RNMT, while there is still a large margin between this transition model and the state-of-the-art model, e.g. the Transformer~\cite{VaswaniEtal2017}, in terms of BLEU, which is also confirmed by~\citeauthor{tangEMNLP18}~\shortcite{tangEMNLP18}.

In this paper, we further enhance the RNMT through increasing the transition depth of the consecutive hidden states along the sequential axis and build a novel and effective Deep Transition RNN-based Architecture for Neural Machine Translation, named \textsc{DTMT}. We design three deep transition modules, which correspondingly extend the RNN modules of shallow RNMT in the encoder and the decoder, to enhance the non-linear transformation between consecutive hidden states. Since the deep transition increases the number of nonlinear steps, this may lead to the problem of vanishing gradients. To alleviate this problem, we propose a Linear Transformation enhanced Gated Recurrent Unit (L-GRU) for \textsc{DTMT}, which provides a linear transformation path throughout the deep transition.

We test the effectiveness of our \textsc{DTMT} on Chinese$\Rightarrow$English, English$\Rightarrow$German and English$\Rightarrow$French translation tasks. Experimental results on NIST Chinese$\Rightarrow$English translation show that \textsc{DTMT} can outperform the Transformer model by +2.09 BLEU points and achieve the best results ever reported in the same dataset. On WMT14 English$\Rightarrow$German and English$\Rightarrow$French translation, it consistently leads to substantial improvements and shows superior quality to the state-of-the-art NMT systems~\cite{VaswaniEtal2017,ChengACL2018}. The main contributions of this paper can be summarized as follows:
\begin{itemize}
\item We tap the potential strength of deep transition between consecutive hidden states and propose a novel deep transition RNN-based architecture for NMT, which achieves state-of-the-art results on multiple translation tasks. \vspace{-2pt}
\item We propose a simple yet more effective linear transformation enhanced GRU for our deep transition RNMT, which provides a linear transformation path for deep transition of consecutive hidden states. Additionally, L-GRU can also be used to enhance other GRU-based architectures, such as the shallow RNMT and the stacked RNMT. \vspace{-2pt}
\item We apply recent advanced techniques, including multi-head attention, layer normalization, label smoothing, and dropouts to enhance our \textsc{DTMT}. Additionally, we find the positional encoding~\cite{VaswaniEtal2017} can assist the training of RNMT by modeling positions of the tokens in the sequence, although it is originally designed for the non-recurrent (self-attentional) architecture.
\end{itemize}

\section{Background} \label{back}
\subsection{Attention-based RNMT}
Given a source sentence $\mathbf{x} \hspace{-3pt}=\hspace{-3pt} \{x_1,x_2,\cdots, x_n\}$ and a target sentence $\mathbf{y} \hspace{-3pt}=\hspace{-3pt} \{y_1,y_2,\cdots, y_m\}$, RNN-based neural machine translation (RNMT) models the translation probability word by word:
\begin{eqnarray}
p(\mathbf{y}|\mathbf{x}) &=& \prod_{t=1}^{m}{P(y_{t}|\mathbf{y_{<t}}, \mathbf{x}; \theta)} \nonumber \\
                         &=& \prod_{t=1}^{m}{softmax(f(\mathbf{c}_t, y_{t-1}, \mathbf{s}_t))} \label{predict}
\end{eqnarray}
where $f(\cdot)$ is a non-linear function, and $\mathbf{s}_t$ is the hidden state of decoder RNN at time step $t$:
\begin{eqnarray}
\mathbf{s}_t = g(\mathbf{s}_{t-1}, y_{t-1}, \mathbf{c}_t) \label{dec_state_update}
\end{eqnarray}
$\mathbf{c}_t$ is a distinct source representation for time $t$, calculated as a weighted sum of the source annotations:
\begin{eqnarray}
\mathbf{c}_t = \sum_{j=1}^{n}{a_{t,j} \mathbf{h}_j} \label{attention_softmax}
\end{eqnarray}
Formally, $\mathbf{h}_j=[\overrightarrow{\mathbf{h}}_j, \overleftarrow{\mathbf{h}}_j]$ is the annotation of $x_j$, which can be computed by a bi-directional RNN~\cite{schuster1997bidirectional} with GRU and contains information about the whole source sentence with a strong focus on the parts surrounding $x_j$. Here,
\begin{eqnarray}
\overrightarrow{\mathbf{h}}_j=\mathbf{GRU}(x_{j}, \overrightarrow{\mathbf{h}}_{j-1}); \hspace{10pt} \overleftarrow{\mathbf{h}}_j=\mathbf{GRU}(x_{j}, \overleftarrow{\mathbf{h}}_{j+1})
\end{eqnarray}
The weight $a_{t,j}$ is computed as
\begin{eqnarray}
a_{t,j} = \frac{exp(e_{t,j})}{\sum_{k=1}^{N}{exp{(e_{t,k})}}} \label{match_score}
\end{eqnarray}
where $e_{t,j}=\mathbf{v}_a^Ttanh(\mathbf{W}_a \tilde{s}_{t-1} + \mathbf{U}_a  \mathbf{h}_j)$ scores how much $\tilde{\mathbf{s}}_{t-1}$ attends to $\mathbf{h}_j$, where $\tilde{\mathbf{s}}_{t-1}=g(\mathbf{s}_{t-1}, y_{t-1})$ is an intermediate state tailored for computing the attention score.

\begin{figure}[t!]
\begin{center}
      \includegraphics[width=0.25\textwidth]{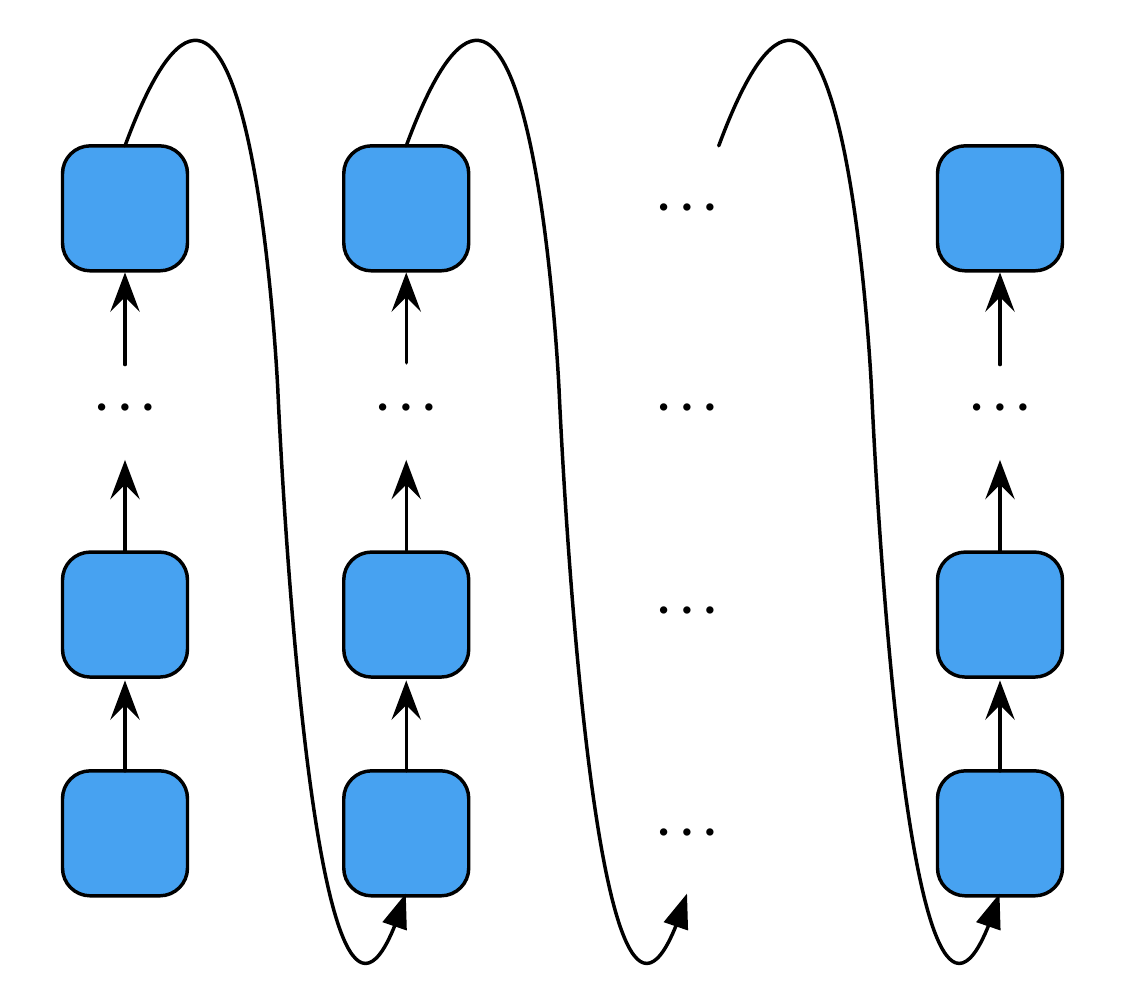}
      \caption{Deep transition RNN, in which the transition between consecutive hidden states is deep.} \label{f:ori_transition}
 \end{center}
\end{figure}

\begin{figure*}[t!]
\begin{center}
      \includegraphics[width=0.9\textwidth]{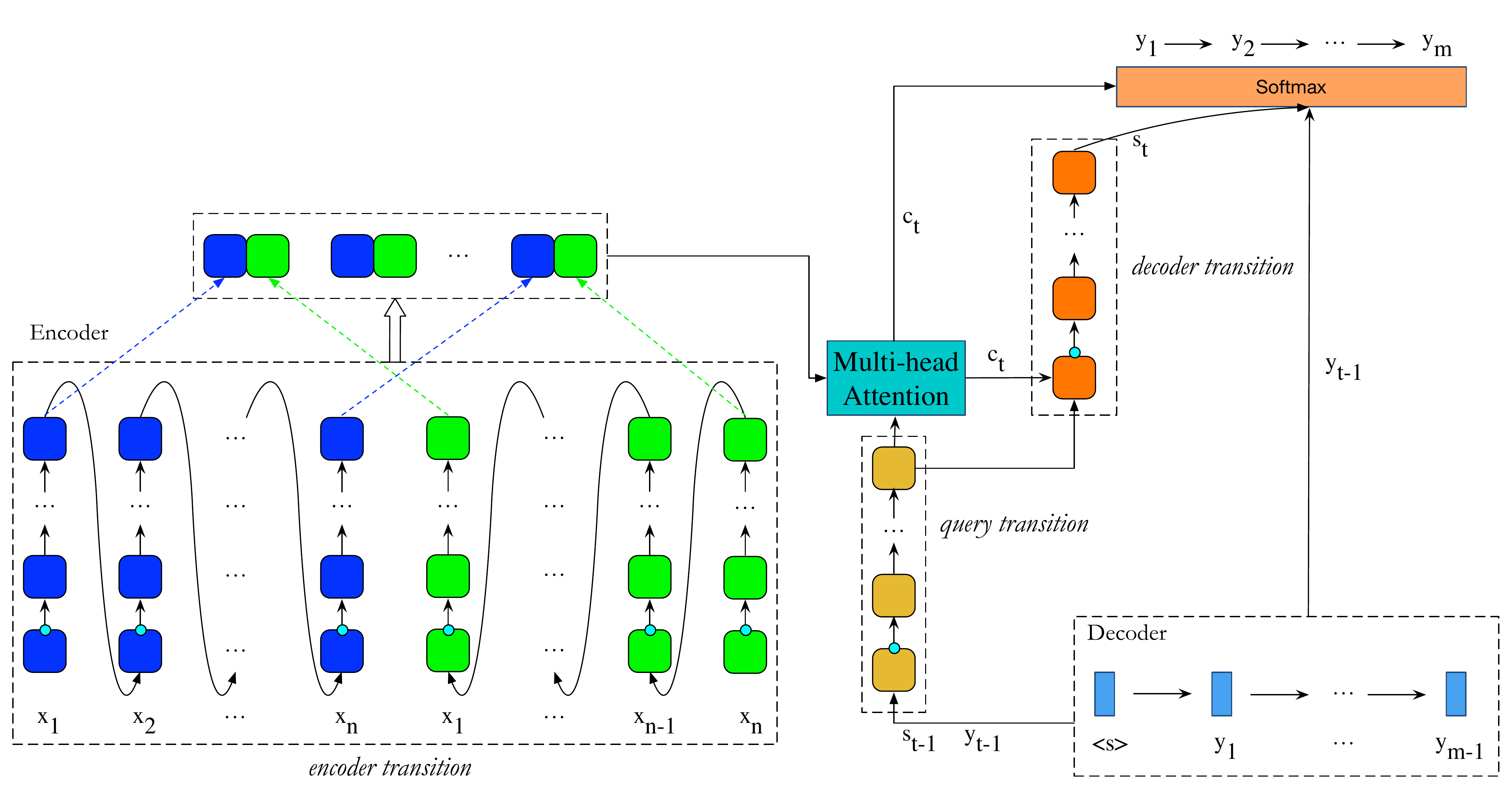}
      \caption{The architecture of \textsc{DTMT}. The bidirectional deep transition encoder (on the left) and the deep transition decoder (on the right) are connected by multi-head attention. There are three deep transition modules, namely the \emph{encoder transition}, the \emph{query transition} and the \emph{decoder transition},  each of which consists of a L-GRU (the square frames fused with a small circle) at the bottom followed by several T-GRUs (the square frames) from bottom to up. } \label{f:deeptransition}
 \end{center}
\end{figure*}

\subsection{Deep Transition RNN}
\citeauthor{barone2017deep}~\shortcite{barone2017deep} first apply the deep transition RNN to NMT.  As shown in Figure~\ref{f:ori_transition}, in a deep transition RNN, the next state is computed by the sequential application of multiple transition layers at each time step, effectively using a feed-forward network embedded inside the recurrent cell. Obviously, this kind of architecture increases the depth of transition between the consecutive hidden states along the sequential axis, unlike the deep stacked RNN, in which transition between the consecutive hidden states is still shallow.

Although the deep transition RNN has been proven to be superior to deep stacked RNN on language modeling task~\cite{pascanu2013construct}, there is still a large margin between this deep transition NMT model~\cite{barone2017deep} and the state-of-the-art NMT model, e.g. the Transformer~\cite{VaswaniEtal2017}, in terms of BLEU, which is also confirmed by~\citeauthor{tangEMNLP18}~\shortcite{tangEMNLP18}.

\section{Model Description} \label{approach}
In this section, we describe our novel Deep Transition RNN-based Architecture for NMT (\textsc{DTMT}). As shown in Figure~\ref{f:deeptransition}, the \textsc{DTMT} consists of a bidirectional deep transition encoder and a deep transition decoder, connected by the multi-head attention~\cite{VaswaniEtal2017}. There are three deep transition modules: 1) \emph{encoder transition} for encoding the source sentence into a sequence of distributed representations; 2) \emph{query transition} for forming a query state to attend to the source representations;  and 3) \emph{decoder transition} for generating the final decoder state of current time step. In each transition module, the transition block consists of a Linear Transformation enhanced GRU (L-GRU) at the bottom followed by several Transition GRUs (T-GRUs) from bottom to up. Before proceeding to the details of \textsc{DTMT}, we first describe the key components (i.e. GRU and its variants) of our deep transition modules.

\subsection{Gated Recurrent Unit and its Variants} \label{grus}
\paragraph{GRU:}
Gated Recurrent Unit (GRU)~\cite{ChoEMNLP} is a variation of LSTM with fewer parameters. The activation function is armed with two specifically designed gates, named update gate and reset gate, to control the flow of information inside each unit. Each hidden state at time-step $t$ is computed as follows:
\begin{eqnarray}
\mathbf{h}_{t} = (1 - \mathbf{z}_t) \odot \mathbf{h}_{t-1} + \mathbf{z}_t \odot \mathbf{\widetilde{h}}_{t}
\end{eqnarray}
where $\odot$ is an element-wise product, $\mathbf{z}_t$ is the update gate, and $\mathbf{\widetilde{h}}_{t}$ is the candidate activation, computed as:
\begin{eqnarray}
\mathbf{\widetilde{h}}_{t} = tanh(\mathbf{W}_{xh}\mathbf{x}_t+\mathbf{r}_t \odot (\mathbf{W}_{hh}\mathbf{h}_{t-1})) \label{activate}
\end{eqnarray}
where $\mathbf{x}_t$ is the input embedding, and $\mathbf{r}_t$ is the reset gate. Reset and update gates are computed as:
\begin{eqnarray}
\mathbf{r}_t = \sigma(\mathbf{W}_{xr}\mathbf{x}_t + \mathbf{W}_{hr}\mathbf{h}_{t-1}) \label{reset} \\
\mathbf{z}_t = \sigma(\mathbf{W}_{xz}\mathbf{x}_t + \mathbf{W}_{hz}\mathbf{h}_{t-1}) \label{update}
\end{eqnarray}
Actually, GRU can be viewed as a non-linear activation function with a specially designed gating mechanism, since the updated $\mathbf{h}_{t}$ has two sources controlled by the update gate and the reset gate: 1) the direct transfer from previous state $\mathbf{h}_{t-1}$; and 2) the candidate update $\mathbf{\widetilde{h}}_{t}$, which is a nonlinear transformation of the previous state $\mathbf{h}_{t-1}$ and the input embedding.

\paragraph{T-GRU:}
Transition GRU (T-GRU) is a key component of deep transition block. A basic deep transition block can be composed of a GRU followed by several T-GRUs from bottom to up at each time step, just as Figure~\ref{f:ori_transition}. In the whole recurrent procedure, for the current time step, the ``state" output of one GRU/T-GRU is used as the ``state" input of the next T-GRU. And the ``state" output of the last T-GRU for the current time step is carried over as the ``state" input of the first GRU for the next time step. For a T-GRU, each hidden state at time-step $t$ is computed as follows:
\begin{eqnarray}
\mathbf{h}_{t} &=& (1 - \mathbf{z}_t) \odot \mathbf{h}_{t-1} + \mathbf{z}_t \odot \mathbf{\widetilde{h}}_{t} \\
\mathbf{\widetilde{h}}_{t} &=& tanh(\mathbf{r}_t \odot (\mathbf{W}_{hh}\mathbf{h}_{t-1}))
\end{eqnarray}
where reset gate $\mathbf{r}_t$ and update gate $\mathbf{z}_t$ are computed as:
\begin{eqnarray}
\mathbf{r}_t = \sigma(\mathbf{W}_{hr}\mathbf{h}_{t-1}) \\
\mathbf{z}_t = \sigma(\mathbf{W}_{hz}\mathbf{h}_{t-1})
\end{eqnarray}
As we can see, T-GRU is a special case of GRU with only ``state" as input. It is also like the convolutional GRU~\cite{KaiserICLR16}. Here the updated $\mathbf{h}_{t}$ has two sources controlled by the update gate and the reset gate: 1) the direct transfer from previous hidden state $\mathbf{h}_{t-1}$; and 2) the candidate update $\mathbf{\widetilde{h}}_{t}$, which is a nonlinear transformation of the previous hidden state $\mathbf{h}_{t-1}$. That is to say, T-GRU conducts both non-linear transformation and direct transfer of the input. This architecture will make training deep models easier.

\paragraph{L-GRU:}
L-GRU is a Linear Transformation enhanced GRU by incorporating an additional linear transformation of the input in its dynamics. Each hidden state at time-step $t$ is computed as follows:
\begin{eqnarray}
\mathbf{h}_{t} = (1 - \mathbf{z}_t) \odot \mathbf{h}_{t-1} + \mathbf{z}_t \odot \mathbf{\widetilde{h}}_{t}
\end{eqnarray}
where the candidate activation $\mathbf{\widetilde{h}}_{t}$ is computed as:
\begin{eqnarray}
\mathbf{\widetilde{h}}_{t} = tanh(\mathbf{W}_{xh}\mathbf{x}_t+\mathbf{r}_t \odot (\mathbf{W}_{hh}\mathbf{h}_{t-1})) + \mathbf{l}_t \odot \mathbf{H}(\mathbf{x}_t) \label{activate}
\end{eqnarray}
where reset gate $\mathbf{r}_t$ and update gate $\mathbf{z}_t$ are computed as the formula (\ref{reset}) and (\ref{update}), and $\mathbf{H}(\mathbf{x}_t) = \mathbf{W}_{x}\mathbf{x}_t$ is the linear transformation of the input $\mathbf{x}_t$, controlled by the linear transformation gate $\mathbf{l}_t$, which is computed as:
\begin{eqnarray}
\mathbf{l}_t = \sigma(\mathbf{W}_{xl}\mathbf{x}_t + \mathbf{W}_{hl}\mathbf{h}_{t-1})
\end{eqnarray}
In L-GRU, the updated $\mathbf{h}_{t}$ has three sources controlled by the update gate, the reset gate and the linear transformation gate: 1) the direct transfer from previous state $\mathbf{h}_{t-1}$; 2) the candidate update $\mathbf{\widetilde{h}}_{t}$; and 3) a direct contribution from the linear transformation of input $\mathbf{H}(\mathbf{x}_t)$. Compared with GRU, L-GRU conducts both non-linear transformation and linear transformation for the inputs, including the embedding input and the state input. Clearly, with L-GRU and T-GRUs, deep transition model can alleviate the problem of vanishing gradients since this structure provides a linear transformation path as a supplement between consecutive hidden states, which are originally connected by only non-linear transformations with multi-steps (e.g. GRU+T-GRUs).

Our L-GRU is inspired by the Linear Associative Unit (LAU)~\cite{wangEtAl2017}, while we exploit more concise operations with the same parameter quantity to the LAU to incorporate the linear transformation of the input $x_t$ as well as preserving the original non-linear abstraction produced by the input and previous hidden state. Different from the LAU, 1) the linear transformation of input is controlled by both the update gate $\mathbf{z}_t$ and the linear transformation gate $\mathbf{l}_t$; and 2) the linear transformation gate $\mathbf{l}_t$ only focus on the linear transformation of the embedding input. These may be the main reasons why L-GRU is more effective than the LAU, as verified in our experiments described later.

\subsection{\textsc{DTMT}}
The formal description of the encoder and the decoder of \textsc{DTMT} is as follows:

\paragraph{Encoder:}
The encoder is a bidirectional deep transition encoder based on recurrent neural networks. Let $L_s$ be the depth of \emph{encoder transition}, then for the $j$th source word in the forward direction the forward source state $\overrightarrow{\mathbf{h}}_j \equiv \overrightarrow{\mathbf{h}}_{j, L_s}$ is computed as:
\begin{eqnarray*}
\overrightarrow{\mathbf{h}}_{j,0} & = & \mathbf{L\textrm{-}GRU}(\mathbf{x}_{j}, \overrightarrow{\mathbf{h}}_{j-1, L_s})  \\
\overrightarrow{\mathbf{h}}_{j,k} & = & \mathbf{T\textrm{-}GRU_k}(\overrightarrow{\mathbf{h}}_{j, k-1})  ~~~~~ \textrm{for}~~1 \leq k \leq L_s
\end{eqnarray*}
where the input to the first L-GRU is the word embedding $\mathbf{x}_j$, while the T-GRUs have only ``state" as the input. Recurrence occurs as the previous state $\overrightarrow{\mathbf{h}}_{j-1, L_s}$ enters the computation in the first L-GRU transition for the current step. The reverse source word states are computed similarly and concatenated to the forward ones to form the bidirectional source annotations $C \equiv \{[\overrightarrow{\mathbf{h}}_{j, L_s}, \overleftarrow{\mathbf{h}}_{j,L_s}]\}$.

\paragraph{Decoder:}
As shown in Figure~\ref{f:deeptransition}, the deep transition decoder consists of two transition modules, named \emph{query transition}  and \emph{decoder transition}, of which \emph{query transition} is conducted before the multi-head attention and \emph{decoder transition} is conducted after the multi-head attention. These transition modules can be conducted to an arbitrary transition depth. Suppose the depth of \emph{query transition} is $L_q$ and the depth of \emph{decoder transition} is $L_d$, then
\begin{eqnarray*}
\mathbf{s}_{t,0} & = & \mathbf{L\textrm{-}GRU}(\mathbf{y}_{t-1}, \mathbf{s}_{t-1, L_q+L_d+1})  \\
\mathbf{s}_{t,k} & = & \mathbf{T\textrm{-}GRU}(\mathbf{s}_{t, k-1})  ~~~~~ \textrm{for}~~1 \leq k \leq L_q
\end{eqnarray*}
where $\mathbf{y}_{t-1}$ is the embedding of the previous target word. And then the context representation $\mathbf{c}_t$ of source sentence is computed by multi-head additive attention:
\begin{eqnarray*}
\mathbf{c}_t = Multihead\textrm{-}Attention(C, \mathbf{s}_{t,L_q})
\end{eqnarray*}
after that, the \emph{decoder transition} is computed as
\begin{eqnarray*}
\mathbf{s}_{t,L_q+1} & = & \mathbf{L\textrm{-}GRU}(\mathbf{c}_{t}, \mathbf{s}_{t,L_q})  \\
\mathbf{s}_{t,L_q+p} & = & \mathbf{T\textrm{-}GRU}(\mathbf{s}_{t, L_q+p-1})  ~~~~~ \textrm{for}~~2 \leq p \leq L_d+1
\end{eqnarray*}

The current state vector $\mathbf{s}_t  \equiv \mathbf{s}_{t,L_q+L_d+1}$ is then used by the feed-forward output network as Formula (\ref{predict}) to predict the current target word.

\begin{table*}[t!]
\centering
\renewcommand{\arraystretch}{1.1}
\scalebox{0.9}{
\begin{tabular}{l| l| c | c | ccccc | l}
\bf \textsc{System}	& \bf \textsc{Architecture} &\bf \# Para. & \bf \textsc{MT06} & \bf \textsc{MT02} & \bf \textsc{MT03} & \bf \textsc{MT04} & \bf \textsc{MT05} & \bf \textsc{MT08} & \bf \textsc{Ave.} \\
\hline
\multicolumn{9}{c}{\em Existing end-to-end NMT systems} \\
\hline
\citeauthor{Shen:15}~\shortcite{Shen:15} 					& GRU with MRT  				& --  		 & 37.34  & 40.36 	& 40.93 & 41.37 & 38.81	& 29.23  & 38.14 \\
\citeauthor{wangEtAl2017}~\shortcite{wangEtAl2017} 		& DeepLAU (4 layers)			& --  		 & 37.29  & -- 		& 39.35 & 41.15 & 38.07	& -- 	      & -- \\
\citeauthor{ZhangAAAI2018}~\shortcite{ZhangAAAI2018}		& Bi-directional decoding			& --  		 & 38.38  & -- 		& 40.02 & 42.32 & 38.84	& --         & -- \\
\citeauthor{MENGIJCAI2018}~\shortcite{MENGIJCAI2018}	& GRU with KV-Memory			&--		 &39.08   & 40.67 	& 38.40 & 41.10 & 38.73	& 30.87  & 37.95 \\
\citeauthor{ChengACL2018}~\shortcite{ChengACL2018} 		& AST (2 layers)    			& --		 & 44.44  & 46.10 	& 44.07 & 45.61 & 44.06   & 34.94  & 42.96 \\
\citeauthor{VaswaniEtal2017}~\shortcite{VaswaniEtal2017} 	& Transformer (\textsc{Big})		& 277.6M	 & 44.78  & 45.32 	& 44.13 & 45.92 & 44.06   & 35.33  & 42.95 \\
\hline
\multicolumn{9}{c}{\em Our end-to-end NMT systems} \\
\hline
\multirow{3}{*}{\em this work}  			&   \textsc{ShallowRNMT} 	& 143.2M & 42.99  & 44.24  & 42.96  & 44.97  & 42.69  & 33.00 & 41.57  \\
   &   \textsc{DTMT\#1} 							& 170.5M & 45.99  & 46.90  & 45.85  & 46.78  & 45.96  & 36.58 & {44.41}\\
   &   \textsc{DTMT\#4} 							& 208.4M & {\bf 46.74}  & {\bf 47.03}  & {\bf 46.34}  & {\bf 47.52}  & {\bf 46.70}  & {\bf 37.61} & {\bf 45.04}\\

\end{tabular}
}
\caption{\label{t:main-result-ch2en} Case-insensitive BLEU scores (\%) on NIST Chinese$\Rightarrow$English translation. Our deep transition model outperforms the state-of-the-art models including the Transformer~\cite{VaswaniEtal2017} and the deep stacked RNMT~\cite{ChengACL2018}.}
\end{table*}

\subsection{Advanced Techniques} \label{techniques}
Except for the multi-head attention, we apply most recently advanced techniques during training to enhance our model:
\begin{itemize}
\item {\bf Dropout:} We apply dropout on embedding layers, the output layer before prediction, and the candidate activation output~\cite{SemeniutaEtal16} of RNN. \vspace{-1pt}
\item {\bf Label Smoothing:} We use uniform label smoothing with an uncertainty=0.1~\cite{SzegedyVISW15}, which has been proved to have a positive impact for the performance.  \vspace{-1pt}
\item {\bf Layer Normalization:} Inspired by the Transformer, per-gate layer normalization~\cite{ba2016layer} is applied within each gate (i.e. reset gate, update gate and linear transformation gate) of L-GRU/T-GRU. It is critical to stabilize the training process of deep transition model.  \vspace{-1pt}
\item {\bf Positional Encoding:} We also add the positional encoding~\cite{VaswaniEtal2017} to the input embeddings at the bottoms of the encoder and decoder to assist modeling positions of the tokens in the sequence. Although the positional encoding is originally designed for the non-recurrent (self-attentional) architecture, we find RNMT can also benefit from it. To stabilize the training process of our deep transition model, we add a scaling factor $1/\sqrt{d_{k}}$ ($d_k$ is the dimension of embedding) to the original positional encoding function.
\end{itemize}

\section{Experiments}\label{experiments}

\subsection{Setup} \label{setup}
We carry out experiments on Chinese$\Rightarrow$English (Zh$\Rightarrow$En), English$\Rightarrow$German (En$\Rightarrow$De) and English$\Rightarrow$French (En$\Rightarrow$Fr) translation tasks. For these tasks, we tokenize the references and evaluated the translation quality with BLEU scores~\cite{papineni2002bleu} as calculated by the \emph{multi-bleu.pl} script.

For Zh$\Rightarrow$En, the training data consists of 1.25M sentence pairs extracted from the LDC corpora. We choose NIST 2006 (MT06) dataset as our valid set, and NIST 2002 (MT02), 2003 (MT03), 2004 (MT04), 2005 (MT05) and 2008 (MT08) datasets as our test sets.
For En$\Rightarrow$De and En$\Rightarrow$Fr, we perform our experiments on the corpora provided by WMT14 that comprise 4.5M and 36M sentence pairs, respectively. We use newstest2013 as the valid set, and newstest2014 as the test set.

\subsection{Training Details} \label{training}
In training the neural networks, we follow~\citeauthor{sennrichACL2016}~\shortcite{sennrichACL2016} to split words into sub-word units. For Zh$\Rightarrow$En, the number of merge operations in byte pair encoding (BPE) is set to 30K for both source and target languages. For En$\Rightarrow$De and En$\Rightarrow$Fr, we use a shared vocabulary generated by 32K BPEs following~\citeauthor{ChenACL2018}~\shortcite{ChenACL2018}.

The parameters are initialized uniformly between [-0.08, 0.08] and updated by SGD with the learning rate controlled by the Adam optimizer~\cite{KingmaB14} ($\beta_1=0.9$, $\beta_2=0.999$, and $\epsilon=1e^{-6}$). And we follow~\citeauthor{ChenACL2018}~\shortcite{ChenACL2018} to vary the learning rate as follows:
\begin{eqnarray}
lr=lr_0 \cdot min(1+t \cdot (n-1) / {np},~n,~n \cdot (2n)^{\frac{s-nt}{e-s}}) \label{learning_decay}
\end{eqnarray}
Here, $t$ is the current step, $n$ is the number of concurrent model replicas in training, $p$ is the number of warmup steps, $s$ is the start step of the exponential decay, and $e$ is the end step of the decay. For Zh$\Rightarrow$En, we use 2 M40 GPUs for synchronous training and set $lr_0$, $p$, $s$ and $e$ to $10^{-3}$, 500, 8000, and 64000 respectively. For En$\Rightarrow$De, we use 8 M40 GPUs and set $lr_0$, $p$, $s$ and $e$ to $10^{-4}$, 50, 200000, and 1200000 respectively. For En$\Rightarrow$Fr, we use 8 M40 GPUs and set $lr_0$, $p$, $s$ and $e$ to $10^{-4}$, 50, 400000, and 3000000 respectively.

We limit the length of sentences to 128 sub-words for Zh$\Rightarrow$En and 256 sub-words for En$\Rightarrow$De and En$\Rightarrow$Fr in the training stage. We batch sentence pairs according to the approximate length, and limit input and output tokens to 4096 per GPU. We apply dropout strategy to avoid over-fitting~\cite{hinton2012improving}. In particular, for Zh$\Rightarrow$En, we set dropout rates of the embedding layers, the layer before prediction and the RNN output layer to 0.5, 0.5 and 0.3 respectively. For En$\Rightarrow$De, we set these dropout rates to 0.3, 0.3 and 0.1 respectively. For En$\Rightarrow$Fr, we set these dropout rates to 0.2, 0.2 and 0.1 respectively. For each model of the translation tasks, the dimension of word embeddings and hidden layer is 1024. Translations are generated by beam search and log-likelihood scores are normalized by the sentence length. We set $beam\_size=4$ and length penalty $alpha=0.6$. We monitor the training process every 2K iterations and decide the early stop condition by validation BLEU.

\subsection{System Description}
\begin{itemize}
\item \textsc{ShallowRNMT}: a shallow yet strong RNMT baseline system, which is our in-house implementation of the attention-based RNMT~\cite{cho} augmented by combining advanced techniques, including multi-head attention, layer normalization, label smoothing, dropouts on multi-layers (the embedding layers, the output layer before prediction and the candidate activation output of each GRU) and the positional encoding.
\item \textsc{DTMT\#num}: our deep transition systems, and the \textsc{\#num} stands for the transition depth  (i.e. the number of T-GRUs) above the bottom L-GRU in each transition module (i.e. \emph{encoder transition}, \emph{query transition}, and \emph{decoder transition}). For example, \textsc{DTMT\#2} means each transition module contains 1 L-GRU layer and 2 T-GRU layers, namely 3 layers in the encoder and 6 layers in the decoder.
\end{itemize}

\begin{table}[t!]
\centering
\scalebox{0.9}{
\begin{tabular}{l| l| l | l }
\bf \textsc{System} & \bf \textsc{Architecture} & \bf \textsc{En-De}  & \bf \textsc{En-Fr}\\
\hline
\citeauthor{ZhouCWLX16}~\shortcite{ZhouCWLX16}		         & LSTM (8 layers)							&20.60 & 37.70 \\
\citeauthor{Luong:15}~\shortcite{Luong:15} 			         & LSTM (4 layers) 							&20.90 & 31.50 \\
\citeauthor{wangEtAl2017}~\shortcite{wangEtAl2017} 		     & DeepLAU (4 layers) 						&23.80 & 35.10 \\
\citeauthor{Wu:16}~\shortcite{Wu:16} 			             & GNMT (8 layers) 							&24.60 & 38.95 \\
\citeauthor{Gehring:17}~\shortcite{Gehring:17} 		         & ConvS2S (15 layers) 						&25.16 & 40.46 \\
\citeauthor{ChengACL2018}~\shortcite{ChengACL2018} 	         & AST (2 layers)        					&25.26 &  -- \\
\citeauthor{VaswaniEtal2017}~\shortcite{VaswaniEtal2017} 	 & Transformer (\textsc{Big})	 			&28.40 & 41.00 \\
\citeauthor{ChenACL2018}~\shortcite{ChenACL2018}		     & RNMT+  (8 layers) 						&28.49 & 41.00 \\
\hline
\multirow{3}{*}{\em this work}  		&   \textsc{ShallowRNMT}    & 25.66 &  39.28 \\
   							&   \textsc{DTMT\#1}  	   & {27.92} & {40.75} \\
   							&   \textsc{DTMT\#4}   	   & {\bf 28.70} & {\bf 42.02}
\end{tabular}
}
\caption{Case-sensitive BLEU scores (\%) on WMT 14 English$\Rightarrow$German and English$\Rightarrow$French translation. \textsc{DTMT\#4} outperforms the state-of-the-art models including the Transformer~\cite{VaswaniEtal2017} and the RNMT+~\cite{ChenACL2018}.}
\label{t:main-result-en2de}
\end{table}

\subsection{Results on NIST Chinese$\Rightarrow$English}
Table~\ref{t:main-result-ch2en} shows the results on NIST Zh$\Rightarrow$En translation task. Our baseline system \textsc{ShallowRNMT} significantly outperforms previous RNN-based NMT systems on the same datasets. \citeauthor{Shen:15}~\shortcite{Shen:15} propose minimum risk training (MRT) to optimize the model with respect to BLEU scores. \citeauthor{wangEtAl2017}~\shortcite{wangEtAl2017} propose the linear associative units (LAU) to address the issue of gradient diffusion, and their system is a deep model with 4 layers. \citeauthor{ZhangAAAI2018}~\shortcite{ZhangAAAI2018} propose to exploit both left-to-right and right-to-left decoding strategies to capture bidirectional dependencies. \citeauthor{MENGIJCAI2018}~\shortcite{MENGIJCAI2018} propose key-value memory augmented attention to improve the adequacy of translation. Compared with them, our baseline system \textsc{ShallowRNMT} outperforms their best models by more than 3 BLEU points. \textsc{ShallowRNMT} is only 1.4 BLEU points lower than the state-of-the-art deep models, i.e. the Transformer (with 6 attention layers)~\cite{VaswaniEtal2017} and the deep stacked RNMT augmented with adversarial stability training (AST)~\cite{ChengACL2018}. We build this strong baseline system to show that the shallow RNMT model with advanced techniques is indeed powerful. And we hope that the strong baseline system used in this work makes the evaluation convincing.

Our deep transition model \textsc{DTMT\#1} with only one transition layer can further bring in up to +3.58 BLEU points (+2.84 BLEU on average) improvements over the strong baseline \textsc{ShallowRNMT}. \textsc{DTMT\#1} outperforms all the previous systems list in Table~\ref{t:main-result-ch2en}, and achieves about +1.45 BLEU points improvements over the best model. With a deeper transition architecture,  \textsc{DTMT\#4} achieves the best results, which is up to +4.61 BLEU points (+3.47 BLEU on average) higher than the \textsc{ShallowRNMT}. Compared with the Transformer and deep RNMT augmented with AST, \textsc{DTMT\#4} yields a gain of +2.08 BLEU on average and achieves the best performance ever reported on this dataset.

\subsection{Results on WMT14 En$\Rightarrow$De and En$\Rightarrow$Fr}
To demonstrate that our models work well across different language pairs, we also evaluate our models on the WMT14 benchmarks on En$\Rightarrow$De and En$\Rightarrow$Fr translation tasks, as listed in Table~\ref{t:main-result-en2de}. For comparison, we list existing NMT systems which are trained on the same WMT 14 corpora. Among these systems, the Transformer~\cite{VaswaniEtal2017} represents the best non-recurrent model, and the RNMT+~\cite{ChenACL2018} represents the best RNMT model. Our baseline system \textsc{ShallowRNMT} can achieve better performance than most RNMT systems except for the RNMT+, which demonstrates that \textsc{ShallowRNMT} is also a strong baseline system for both En$\Rightarrow$De and En$\Rightarrow$Fr.

Our deep transition model  \textsc{DTMT\#1} with only one transition layer can bring in +2.26 BLEU for En$\Rightarrow$De and +1.47 BLEU for En$\Rightarrow$Fr over the strong baseline \textsc{ShallowRNMT}. With a deeper transition architecture, our \textsc{DTMT\#4} achieves +3.04 BLEU for En$\Rightarrow$De and +2.74 BLEU for En$\Rightarrow$Fr over the baseline, and outperforms state-of-the-art systems, i.e. the Transformer and the RNMT+.

\subsection{Analysis}
\paragraph{L-GRU vs. GRU \& LAU:}
We investigate the effectiveness of the proposed L-GRU on different architectures, including the \textsc{ShallowRNMT} and the \textsc{DTMT}s. From Table~\ref{t:refined-linear-gate} we can see that the L-GRU is effective since it can consistently bring in substantial improvements over different architectures. In particular, it brings in +2.26 BLEU points improvements over \textsc{ShallowRNMT} averagely on five test sets, and it also leads to +0.78 $\sim$ +0.88 BLEU points improvements over our deep transition architectures (i.e. \textsc{DTMT\#1} and \textsc{DTMT\#4}). Additionally, with more concise operations and the same parameter quantity, L-GRU can further outperform the LAU~\cite{wangEtAl2017} on different strong systems by +0.5 $\sim$ +0.77 BLEU points. These results demonstrate that L-GRU is a more effective unit for both deep transition models and the shallow RNMT model.

\begin{table}[t!]
\centering
\renewcommand{\arraystretch}{1.1}
\scalebox{0.9}{
\begin{tabular}{l| l| c | c}
\bf \textsc{Architecture}	& \bf \textsc{RNN} &\bf \# Para. &  \bf \textsc{BLEU} \\
\hline
\multirow{3}{*}{\textsc{ShallowRNMT}}  		& \textsc{GRU} 						& 143.2M & 41.57  \\
									& \textsc{\emph{LAU}} 				& 157.9M & 43.06  \\
									& \textsc{\emph{L-GRU}} 				& 157.9M & 43.83 \\
\hline
\multirow{3}{*}{\textsc{DTMT\#1}}  			& \textsc{\emph{GRU+T-GRU}} 		& 155.8M & 43.63  \\
									& \textsc{\emph{LAU+T-GRU}} 			& 170.5M & 43.79  \\
									& \textsc{L-GRU+T-GRU} 				& 170.5M & 44.41 \\
\hline
\multirow{3}{*}{\textsc{DTMT\#4}}  			& \textsc{\emph{GRU+T-GRUs}} 		& 193.7M & 44.16 \\
									& \textsc{\emph{LAU+T-GRUs}} 		& 208.4M & 44.54 \\
									& \textsc{L-GRU+T-GRUs} 			& 208.4M & {\bf 45.04} \\

\end{tabular}
}
\caption{\label{t:refined-linear-gate} Comparisons of GRU, LAU and L-GRU with different architectures on NIST Chinese$\Rightarrow$English translation (average BLEU scores (\%) on test sets). The italics in the ``RNN" column indicate the potential variants of the corresponding architecture. For example, the RNN units in \textsc{DTMT\#1} can be replaced with the \textsc{\emph{LAU+T-GRU}}.
}
\end{table}

\paragraph{Ablation Study:}
Our deep transition model consists of three deep transition modules, including the \emph{encoder transition}, the \emph{query transition} and the \emph{decoder transition}. We perform an ablation study on Zh$\Rightarrow$En translation to investigate the effectiveness of these transition modules by choosing \textsc{DTMT\#4} as an example. As shown in Table~\ref{t:transition}, replacing any one with its corresponding part of the shallow RNMT~\cite{cho} leads to the translation performance decrease (-0.68 $\sim$ -1.74 BLEU). Among these transition modules, the \emph{encoder transition} is the most important, since deleting it leads to the most obvious decline (-1.74 BLEU). We also conduct an ablation study of the L-GRU and the ``Advanced Techniques" on En$\Rightarrow$De task. As shown in Table~\ref{t:ablation_lgru_tecs}, deleting the L-GRU and/or the ``Advanced Techniques" leads to sharp declines on translation quality. Therefore, we can conclude that both the L-GRU and the ``Advanced Techniques" are key components for \textsc{DTMT\#4} to achieve the state-of-the-art performance.

\paragraph{Transition Depth \& Positional Encoding:}
Table~\ref{t:depth_and_pe} shows the impact of the transition depth and the positional encoding on Zh$\Rightarrow$En translation. From these results, we can draw the following conclusions: 1) with the increasing of transition depth (rows 3-6), our model can consistently achieve better performance; 2) T-GRUs do bring in significant improvements even over the strong baseline (rows 2-3); and 3) on different architectures, we can see that the positional encoding can further bring in consistent improvements (+0.1 $\sim$ +0.3 BLEU) over its counterpart without the positional encoding. This demonstrates that, although the positional encoding is originally designed for non-recurrent network (i.e. Transformer), it also can assist the training of RNN-based models by modeling positions of tokens in the sequence.

\begin{table}[t!]
\centering
\renewcommand{\arraystretch}{1.1}
\scalebox{0.9}{
\begin{tabular}{c | c | c | l}
\bf $\emph{enc-transion}$ & \bf $\emph{query-transion}$ & \bf $\emph{dec-transion}$ & \bf \textsc{BLEU} \\
\hline
$\surd$ & $\surd$ & $\surd$ 	& {\bf 45.04} \\
$\times$ & $\surd$ & $\surd$  	& 43.30\\
$\surd$ & $\times$ & $\surd$  	& 44.35\\
$\surd$ & $\surd$ & $\times$  	& 44.36\\
\end{tabular}
}
\caption{\label{t:transition} Ablation study of deep transition modules on NIST Chinese$\Rightarrow$English translation (average BLEU scores (\%) on test sets). Here ``$\times$" stands for replacing the transition module with the corresponding part of the conventional shallow RNMT~\cite{cho}.
}
\end{table}

\begin{table}[t!]
\centering
\renewcommand{\arraystretch}{1.1}
\scalebox{0.9}{
\begin{tabular}{l | l}
\bf \textsc{Architecture} & \bf \textsc{BLEU} \\
\hline
\textsc{DTMT\#4} 							&  {\bf 28.70} \\
~~~~$-$\textsc{L-GRU}       					& 27.81\\
~~~~$-$\textsc{L-GRU} \& Advanced Techniques  	& 26.27\\
\end{tabular}
}
\caption{\label{t:ablation_lgru_tecs} Ablation study of the L-GRU and the ``Advanced Techniques" on WMT14 English$\Rightarrow$German translation.}
\end{table}

\begin{figure}[t!]
\begin{center}
      \includegraphics[width=0.4\textwidth]{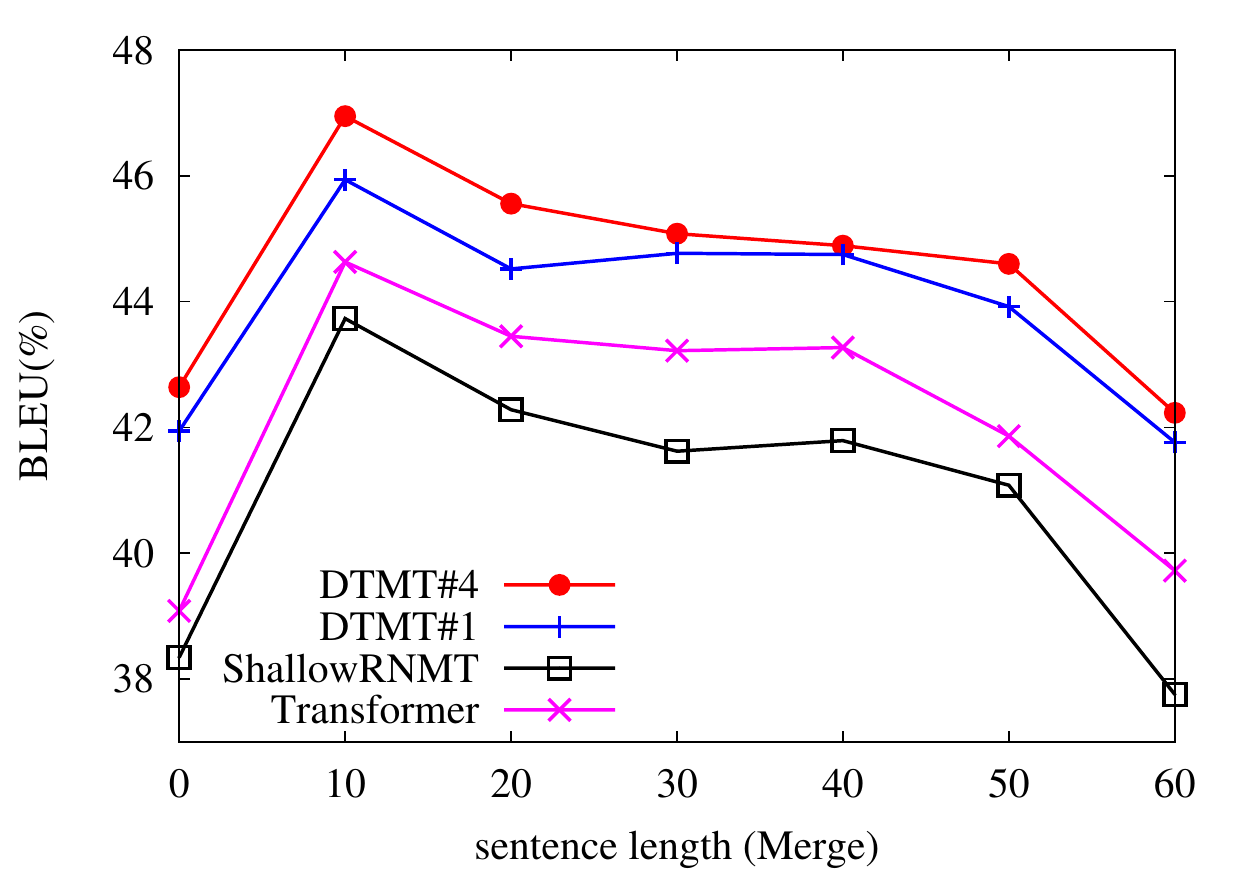} \vspace{-5pt}
      \caption{The BLEU scores (\%) of generated translations on the merged four test sets with respect to the lengths of source sentences. The numbers on X-axis of the figure stand for sentences \emph{longer than} the corresponding length, e.g., $40$ for source sentences with $>40$ words.
} \label{f:res_length}
 \end{center}
\end{figure}

\paragraph{About Length:}
A more detailed comparison between $\textsc{DTMT\#4}$, $\textsc{DTMT\#1}$,  $\textsc{ShallowRNMT}$ and the Transformer suggest that our deep transition architectures are essential to achieve the superior performance. Figure~\ref{f:res_length} shows the BLEU scores of generated translations on the test sets with respect to the lengths of the source sentences. In particular, we test the BLEU scores on sentences longer than \{0, 10, 20, 30, 40, 50, 60\} in the merged test set of MT02, MT03, MT04, MT05 and MT08. Clearly, on sentences with different lengths, $\textsc{DTMT\#4}$ and $\textsc{DTMT\#1}$ always yield higher BLEU scores than \textsc{ShallowRNMT} and the Transformer consistently. And $\textsc{DTMT\#4}$ yields the best BLEU scores on sentences with different lengths.

\begin{table}[t!]
\centering
\renewcommand{\arraystretch}{1.1}
\scalebox{0.9}{
\begin{tabular}{c| l| c| c}
\bf \textsc{\#} & \bf \textsc{Architecture}	& \bf \textsc{Non-PE} & \bf \textsc{PE} \\
\hline
1 & \textsc{ShallowRNMT} 				& 41.22  & 41.57    \\
2 & ~~~~~~~~~~~~~~+ \textsc{L-GRU} 		& 43.54  & 43.83    \\
\hline
3 & \textsc{DTMT\#1} 					& 44.28  & 44.41    \\
4 & \textsc{DTMT\#2} 					& 44.50  & 44.66    \\
5 & \textsc{DTMT\#3} 					& 44.61  & 44.70    \\
6 & \textsc{DTMT\#4} 					& {\bf 44.72}  & {\bf 45.04}    \\
\end{tabular}
}
\caption{\label{t:depth_and_pe} Impact of transition depth and the positional encoding (PE) on NIST Chinese$\Rightarrow$English translation (average BLEU scores (\%) on test sets).
}
\end{table}

\subsection{Related Work}
Our work is inspired by the deep transition RNN~\cite{pascanu2013construct}, which is applied on language modeling task. \citeauthor{barone2017deep}~\shortcite{barone2017deep} fist apply this kind of architecture on NMT, while there is still a large margin between this transition model and the state-of-the-art NMT models. Different from these works, we extremely enhance the deep transition architecture and build the state-of-the-art deep transition NMT model from three aspects: 1) fusing L-GRU and T-GRUs, to provide a linear transformation path between consecutive hidden states, as well as preserving the non-linear transformation path; 2) exploiting three deep transition modules, including the \emph{encoder transition}, the \emph{query transition} and the \emph{decoder transition}; and 3) investigating and combing recent advanced techniques, including multi-head attention, labeling smoothing, layer normalization, dropout on multi-layers and positional encoding.

Our work is also inspired by deep stacked RNN models for NMT~\cite{ZhouCWLX16,wangEtAl2017,ChenACL2018}. \citeauthor{ZhouCWLX16}~\shortcite{ZhouCWLX16} propose fast-forward connections to address the notorious problem of vanishing/exploding gradients for deep stacked RNMT. \citeauthor{wangEtAl2017}~\shortcite{wangEtAl2017} propose the Linear Associative Unit (LAU) to reduce the gradient path inside the recurrent units. Different from these studies, we focus on the deep transition architecture and propose a novel linear transformation enhanced GRU (L-GRU) for our deep transition RNMT. L-GRU is verified more effective than the LAU, although L-GRU exploits more concise operations with the same parameter quantity to incorporate the linear transformation of the embedding input. Inspired by RNMT+~\cite{ChenACL2018}, we investigate and combine generally applicable training and optimization techniques, and finally enable our \textsc{DTMT} to achieve superior quality to state-of-the-art NMT systems.

\subsection{Conclusion}
We propose a novel and effective deep transition architecture for NMT. Our empirical study on Chinese$\Rightarrow$English,  English$\Rightarrow$German and English$\Rightarrow$French translation tasks shows that our DTMT can achieve remarkable improvements on translation quality. Experimental results on NIST Chinese$\Rightarrow$English translation show that \textsc{DTMT} can outperform the Transformer by +2.09 BLEU points even with fewer parameters and achieve the best results ever reported on the same dataset. On WMT14 English$\Rightarrow$German and English$\Rightarrow$French tasks, it shows superior quality to the state-of-the-art  NMT systems~\cite{VaswaniEtal2017,ChenACL2018}.

\bibliographystyle{named}
\bibliography{aaai}

\begin{thebibliography}{}

\bibitem[\protect\citeauthoryear{Ba \bgroup \em et al.\egroup
  }{2016}]{ba2016layer}
Jimmy~Lei Ba, Jamie~Ryan Kiros, and Geoffrey~E Hinton.
\newblock Layer normalization.
\newblock {\em arXiv preprint arXiv:1607.06450}, 2016.

\bibitem[\protect\citeauthoryear{Bahdanau \bgroup \em et al.\egroup
  }{2015}]{cho}
Dzmitry Bahdanau, Kyunghyun Cho, and Yoshua Bengio.
\newblock Neural machine translation by jointly learning to align and
  translate.
\newblock In {\em ICLR}, 2015.

\bibitem[\protect\citeauthoryear{Barone \bgroup \em et al.\egroup
  }{2017}]{barone2017deep}
Antonio Valerio~Miceli Barone, Jind{\v{r}}ich Helcl, Rico Sennrich, Barry
  Haddow, and Alexandra Birch.
\newblock Deep architectures for neural machine translation.
\newblock {\em arXiv preprint arXiv:1707.07631}, 2017.

\bibitem[\protect\citeauthoryear{Chen \bgroup \em et al.\egroup
  }{2018}]{ChenACL2018}
Mia~Xu Chen, Orhan Firat, Ankur Bapna, Melvin Johnson, Wolfgang Macherey,
  George Foster, Llion Jones, Mike Schuster, Noam Shazeer, Niki Parmar, Ashish
  Vaswani, Jakob Uszkoreit, Lukasz Kaiser, Zhifeng Chen, Yonghui Wu, and
  Macduff Hughes.
\newblock The best of both worlds: Combining recent advances in neural machine
  translation.
\newblock In {\em ACL}, 2018.

\bibitem[\protect\citeauthoryear{Cheng \bgroup \em et al.\egroup
  }{2018}]{ChengACL2018}
Yong Cheng, Zhaopeng Tu, Fandong Meng, Junjie Zhai, and Yang Liu.
\newblock Towards robust neural machine translation.
\newblock In {\em ACL}, 2018.

\bibitem[\protect\citeauthoryear{Cho \bgroup \em et al.\egroup
  }{2014}]{ChoEMNLP}
Kyunghyun Cho, Bart van Merrienboer, Caglar Gulcehre, Fethi Bougares, Holger
  Schwenk, and Yoshua Bengio.
\newblock Learning phrase representations using rnn encoder-decoder for
  statistical machine translation.
\newblock In {\em EMNLP}, 2014.

\bibitem[\protect\citeauthoryear{Gehring \bgroup \em et al.\egroup
  }{2017}]{Gehring:17}
Jonas Gehring, Michael Auli, David Grangier, Denis Yarats, and Yann~N Dauphin.
\newblock Convolutional sequence to sequence learning.
\newblock In {\em ICML}, 2017.

\bibitem[\protect\citeauthoryear{He \bgroup \em et al.\egroup
  }{2016}]{he2016deep}
Kaiming He, Xiangyu Zhang, Shaoqing Ren, and Jian Sun.
\newblock Deep residual learning for image recognition.
\newblock In {\em CVPR}, 2016.

\bibitem[\protect\citeauthoryear{Hinton \bgroup \em et al.\egroup
  }{2012}]{hinton2012improving}
Geoffrey~E Hinton, Nitish Srivastava, Alex Krizhevsky, Ilya Sutskever, and
  Ruslan~R Salakhutdinov.
\newblock Improving neural networks by preventing co-adaptation of feature
  detectors.
\newblock {\em arXiv}, 2012.

\bibitem[\protect\citeauthoryear{Kaiser and Sutskever}{2015}]{KaiserICLR16}
Lukasz Kaiser and Ilya Sutskever.
\newblock Neural gpus learn algorithms.
\newblock {\em CoRR}, abs/1511.08228, 2015.

\bibitem[\protect\citeauthoryear{Kalchbrenner \bgroup \em et al.\egroup
  }{2017}]{Kalchbrenner:16}
Nal Kalchbrenner, Lasse Espeholt, Karen Simonyan, Aaron van~den Oord, Alex
  Graves, and Koray Kavukcuoglu.
\newblock Neural machine translation in linear time.
\newblock In {\em ICML}, 2017.

\bibitem[\protect\citeauthoryear{Kingma and Ba}{2014}]{KingmaB14}
Diederik~P. Kingma and Jimmy Ba.
\newblock Adam: {A} method for stochastic optimization.
\newblock {\em CoRR}, abs/1412.6980, 2014.

\bibitem[\protect\citeauthoryear{Luong \bgroup \em et al.\egroup
  }{2015}]{Luong:15}
Minh-Thang Luong, Hieu Pham, and Christopher~D Manning.
\newblock Effective approaches to attention-based neural machine translation.
\newblock In {\em EMNLP}, 2015.

\bibitem[\protect\citeauthoryear{Meng \bgroup \em et al.\egroup
  }{2018}]{MENGIJCAI2018}
Fandong Meng, Zhaopeng Tu, Yong Cheng, Haiyang Wu, Junjie Zhai, Yuekui Yang,
  and Di~Wang.
\newblock Neural machine translation with key-value memory-augmented attention.
\newblock In {\em IJCAI}, 2018.

\bibitem[\protect\citeauthoryear{Papineni \bgroup \em et al.\egroup
  }{2002}]{papineni2002bleu}
Kishore Papineni, Salim Roukos, Todd Ward, and Wei-Jing Zhu.
\newblock Bleu: a method for automatic evaluation of machine translation.
\newblock In {\em ACL}, 2002.

\bibitem[\protect\citeauthoryear{Pascanu \bgroup \em et al.\egroup
  }{2014}]{pascanu2013construct}
Razvan Pascanu, Caglar Gulcehre, Kyunghyun Cho, and Yoshua Bengio.
\newblock How to construct deep recurrent neural networks.
\newblock In {\em ICLR}, 2014.

\bibitem[\protect\citeauthoryear{Schuster and
  Paliwal}{1997}]{schuster1997bidirectional}
Mike Schuster and Kuldip~K Paliwal.
\newblock Bidirectional recurrent neural networks.
\newblock {\em TSP}, 1997.

\bibitem[\protect\citeauthoryear{Semeniuta \bgroup \em et al.\egroup
  }{2016}]{SemeniutaEtal16}
Stanislau Semeniuta, Aliaksei Severyn, and Erhardt Barth.
\newblock Recurrent dropout without memory loss.
\newblock {\em CoRR}, abs/1603.05118, 2016.

\bibitem[\protect\citeauthoryear{Sennrich \bgroup \em et al.\egroup
  }{2016}]{sennrichACL2016}
Rico Sennrich, Barry Haddow, and Alexandra Birch.
\newblock Neural machine translation of rare words with subword units.
\newblock In {\em ACL}, 2016.

\bibitem[\protect\citeauthoryear{Shen \bgroup \em et al.\egroup
  }{2016}]{Shen:15}
Shiqi Shen, Yong Cheng, Zhongjun He, Wei He, Hua Wu, Maosong Sun, and Yang Liu.
\newblock Minimum risk training for neural machine translation.
\newblock In {\em ACL}, 2016.

\bibitem[\protect\citeauthoryear{Sutskever \bgroup \em et al.\egroup
  }{2014}]{googleS2S}
Ilya Sutskever, Oriol Vinyals, and Quoc~VV Le.
\newblock Sequence to sequence learning with neural networks.
\newblock In {\em NIPS}, 2014.

\bibitem[\protect\citeauthoryear{Szegedy \bgroup \em et al.\egroup
  }{2015}]{SzegedyVISW15}
Christian Szegedy, Vincent Vanhoucke, Sergey Ioffe, Jonathon Shlens, and
  Zbigniew Wojna.
\newblock Rethinking the inception architecture for computer vision.
\newblock {\em CoRR}, abs/1512.00567, 2015.

\bibitem[\protect\citeauthoryear{Tang \bgroup \em et al.\egroup
  }{2018}]{tangEMNLP18}
Gongbo Tang, Mathias M{\"u}ller, Annette Rios, and Rico Sennrich.
\newblock {Why Self-Attention? A Targeted Evaluation of Neural Machine
  Translation Architectures}.
\newblock In {\em {EMNLP}}, 2018.

\bibitem[\protect\citeauthoryear{Vaswani \bgroup \em et al.\egroup
  }{2017}]{VaswaniEtal2017}
Ashish Vaswani, Noam Shazeer, Niki Parmar, Jakob Uszkoreit, Llion Jones,
  Aidan~N. Gomez, Lukasz Kaiser, and Illia Polosukhin.
\newblock Attention is all you need.
\newblock In {\em NIPS}, 2017.

\bibitem[\protect\citeauthoryear{Wang \bgroup \em et al.\egroup
  }{2017}]{wangEtAl2017}
Mingxuan Wang, Zhengdong Lu, Jie Zhou, and Qun Liu.
\newblock Deep neural machine translation with linear associative unit.
\newblock In {\em ACL}, 2017.

\bibitem[\protect\citeauthoryear{Wu \bgroup \em et al.\egroup }{2016}]{Wu:16}
Yonghui Wu, Mike Schuster, Zhifeng Chen, Quoc~V Le, Mohammad Norouzi, Wolfgang
  Macherey, Maxim Krikun, et~al.
\newblock Google's neural machine translation system: Bridging the gap between
  human and machine translation.
\newblock {\em arXiv preprint arXiv:1609.08144}, 2016.

\bibitem[\protect\citeauthoryear{Zhang \bgroup \em et al.\egroup
  }{2018}]{ZhangAAAI2018}
Xiangwen Zhang, Jinsong Su, Yue Qin, Yang Liu, Rongrong Ji, and Hongji Wang.
\newblock Asynchronous bidirectional decoding for neural machine translation.
\newblock In {\em AAAI}, 2018.

\bibitem[\protect\citeauthoryear{Zhou \bgroup \em et al.\egroup
  }{2016}]{ZhouCWLX16}
Jie Zhou, Ying Cao, Xuguang Wang, Peng Li, and Wei Xu.
\newblock Deep recurrent models with fast-forward connections for neural
  machine translation.
\newblock {\em TACL}, 2016.

\end{thebibliography}

\end{document}